\documentclass[runningheads]{llncs}
\usepackage[T1]{fontenc}
\usepackage{graphicx}
\usepackage{glossaries}
\usepackage{makecell}
\usepackage{tabularx,booktabs} 
\usepackage{xcolor}
\usepackage{biblatex}
\usepackage{hyperref}
\newacronym{xai}{\textit{XAI}}{\textit{Explainable Artificial Intelligence}}
\newacronym{pg}{\textit{PG}}{\textit{Policy Graph}}
\newacronym{ipg}{\textit{IPG}}{\textit{Intention-aware Policy Graph}}

\begin{document}
\title{Explaining Autonomous Vehicles with Intention-aware Policy Graphs}

\author{Sara Montese\inst{1,2}\orcidID{0009-0000-6446-7871} \and
Victor Gimenez-Abalos \inst{1,2}\orcidID{0000-0003-4514-6145} \and
Atia Cortés \inst{1}\orcidID{0000-0002-4394-439X} \and
Ulises Cortés \inst{2,1}\orcidID{0000-0003-0192-3096}\and
Sergio Alvarez-Napagao \inst{2,1}\orcidID{0000-0001-9946-9703}}

\authorrunning{S. Montese et al.}
\institute{Barcelona Supercomputing Center, Barcelona, Spain \and
Universitat Politècnica de Catalunya, Barcelona, Spain\\
\email{\{sara.montese, victor.gimenez, sergio.alvarez\}@bsc.es}} 
\maketitle              

\begin{abstract}
The potential to improve road safety, reduce human driving error, and promote environmental sustainability have enabled the field of autonomous driving to progress rapidly over recent decades. The performance of autonomous vehicles has significantly improved thanks to advancements in Artificial Intelligence, particularly Deep Learning. Nevertheless, the opacity of their decision-making, rooted in the use of accurate yet complex AI models, has created barriers to their societal trust and regulatory acceptance, raising the need for explainability. We propose a \textit{post-hoc}, model-agnostic solution to provide teleological explanations for the behaviour of an autonomous vehicle in urban environments. 
Building on Intention-aware Policy Graphs, our approach enables the extraction of interpretable and reliable explanations of vehicle behaviour in the nuScenes dataset from global and local perspectives. 
We demonstrate the potential of these explanations to assess whether the vehicle operates within acceptable legal boundaries and to identify possible vulnerabilities in autonomous driving datasets and models.



\end{abstract}
\section{Introduction}
Autonomous driving (AD) has made remarkable strides over the past two decades, propelled by advancements in Artificial Intelligence (AI). 
Despite the promising potential of autonomous vehicles (AVs) to revolutionise daily transportation, their lack of transparency, led by the use of AI, raises concerns in terms of their functional safety, 
accountability, ethical dilemmas
, and regulatory compliance~\cite{Atakishiyev_AVSurvey_2024}. 
Hence, explainability is critical for designing safe and trustworthy AD systems. 

While most explainability methods~\cite{Kuznietsov_AVSurvey_2024} in AD focus on identifying the mechanisms influencing a vehicle’s actions, they often fail to elucidate the reasons or motivations behind such specific decisions. 
Moreover, these explanations are commonly oriented towards individuals with technical expertise (\textit{e.g.} AV developers) rather than being directed to a broader audience~\cite{XAI_not_meeting_expectations}. 
These limitations have led to the development of \gls{xai} techniques inspired by social sciences~\cite{miller2019} that aim to build trust and acceptance of AVs by conveying understandable information to non-technical users.

In this regard, we propose an explainability approach to elicit teleological and intelligible explanations of AV behaviour based on an existing \gls{xai} method called \gls{ipg}~\cite{IPGs}.
We use driving trajectories from the nuScenes dataset~\cite{nuscenes} to build an \gls{ipg} as a probabilistic graphical model that represents the behaviour of the observed vehicle while incorporating folk-psychological notions of desires and intentions~\cite{malle2004}. 
This method serves as a \textit{post-hoc}, model-agnostic solution applicable to any AD system without requiring access to its internal model. 
The explanations generated focus on global and local aspects of the behaviour. They can be delivered using intuitive visualisations and natural language predicates, thereby extending interpretability to users with different levels of expertise.

The contributions of this work are threefold. 
First, we develop a framework that integrates the vehicle's desires (\textit{e.g.} to allow a pedestrian to cross) and associated intentions to extract the motivations underlying both general driving behaviour and specific driving decisions, with the possibility of identifying abnormal or undesirable conduct (\textit{e.g. }not halting at a stop sign). 
The interpretability and reliability of the explanations produced are assessed by computing specific intention metrics over the \gls{ipg}, and then presented via intuitive visualisations.

Second, we investigate how external factors, such as weather conditions and time of day, influence AV decision-making, uncovering potential harmful \textit{biases} and 
vulnerabilities in the dataset. 

Finally, we enable explainees to interact with the vehicle surrogate model by posing telic explainability questions, such as ``What do you intend to do in state \textit{s}?'', ``Why would you do action $a$ in state $s$?'', and ``How do you plan to fulfil intention \textit{I} in state \textit{s}?''.

\section{Background and Related Work}
Several surveys have analysed and categorised different methods for \gls{xai} in AVs~\cite{Kuznietsov_AVSurvey_2024, Atakishiyev_AVSurvey_2024}. 
Most approaches, such as vision-based~\cite{grad_cam_XAV, saliency_XAV} and relevance-based~\cite{shap_XAV} techniques
, view explainability in a solely technical manner and explain a system's actions by identifying the mechanical factors that influenced them without being able to explain the motivations behind such specific decisions. 
In addition, these explanations often rely on visual or numerical representations, which are less intuitive and harder to interpret for users with no technical background (\textit{e.g.} drivers, legislators)~\cite{Kuznietsov_AVSurvey_2024}.
Such technical-centric 
explanations fail to consider the needs and expectations of non-expert users, thereby hindering progress in building trust in AVs~\cite{XAI_not_meeting_expectations}.
This inadequacy has led towards \gls{xai} approaches inspired by social sciences, grounded in \textit{how} humans select 
and present explanations to one another~\cite{miller2019}.

Malle's folk-conceptual framework for behaviour explanation~\cite{malle2004} posits that people provide explanations for other people's actions by attributing them mental states such as beliefs (B), desires (D) and intentions (I) and that for \textit{intentional} actions, explanations often focus on the underlying goals driving the behaviour. Extending this to AVs, research has shown that purpose-oriented explanations are the most satisfying for human users, even when the behaviour has originated from an AI system~\cite{people_attribute_purpose_to_AV}. Interpreting intentions can therefore be key to making opaque AV decision-making in autonomous vehicles more transparent, explainable, and aligned with human values. 

While many existing explainability methods focus on revealing which inputs have influenced an AV's actions, they fail to capture the motivations and reasons behind these choices. Moreover, they lack objective measures to assess foreseeability, making them insufficient to ensure accountability~\cite{kraus_towards_2024}.
As discussed in~\cite{interpreting_intent_is_key}, intent-based explanations provide a structured and human-centric approach to evaluating whether AV decisions align with legal, ethical and safety standards. This is particularly relevant for regulatory compliance, where intent plays a critical role in auditing regulated behaviour (such as AV behaviour) and assigning accountability in case of incidents.

Beyond regulatory concerns, intention-aware modelling should 
enable further coordination capabilities in multi-agent environments, where an AV must infer and respond to the near-future actions of pedestrians, cyclists, and other vehicles. As noted in~\cite{interpreting_intent_is_key}, trust in AI systems improves when they exhibit goal-driven, interpretable behaviour rather than acting as reactive, opaque entities.

In this context, Gimenez-Abalos \textit{et al.}~\cite{IPGs} proposed a \textit{post-hoc}, model-agnostic solution, Intention-aware Policy Graphs, to extract global and local explanations of an agent's behaviour in an intelligible format. 
Here, a \gls{pg} is constructed by observing an agent's behaviour and representing it as a directed graph, where each node corresponds to a state discretised using natural language predicates, and each edge denotes the transition probability between states from the execution of an action available in the environment. 
Previous work~\cite{singleagent-PG, multiagent-PG} demonstrated that \gls{pg}s can be traversed to elicit cause explanations 
about an agent's decisions in single-agent and multi-agent environments. \gls{ipg}s extend \gls{pg}s by incorporating intentions and further enabling the generation of teleological explanations -- that is, explanations that clarify \textit{why} an action was taken in terms of the agent's designed or inferred goals.

With this research, we explore the application of \gls{ipg}s to understand the rationale of autonomous vehicles, as by integrating \gls{ipg}s we provide a framework that formalises and quantifies an AV's inferred intentions, enabling structured counterfactual reasoning, legal verification, and safer human-AI interaction.

Other \gls{xai} methods contribute to alternative human-centred explanations for AVs.
Gyevnar \textit{et al.}~\cite{cema, XAVI} proposed two approaches to elicit causal explanations of a vehicle's decisions in natural language. 
Unlike these methods, our approach does not require \textit{prior} knowledge of specific components of the vehicle's design (\textit{i.e.} the reward function) or make assumptions about its decision-making process. 
This flexibility allows the proposed method to adapt to scenarios where a vehicle might have a variable number of goals or its actions are not exclusively goal-oriented.
 
Omeiza \textit{et al.}~\cite{Omeiza_XAV} generate natural language explanations using tree-based representations of driving scenes; however, this method does not consider the temporal context of driving decisions and the objectives that influence them, thus limiting its applicability to sequential decision-making in dynamic environments.

\section{\gls{ipg}s for AV Behaviour}
We aim to explain the decisions made by an AV (hereafter referred to as \textit{ego vehicle}) navigating within an urban environment.
The explanatory process starts by collecting observations of the ego vehicle's behaviour.
At time $t$, the state of the ego vehicle $s_t\in S$ consists of its position, heading, velocity, acceleration, and steering angle. 
Additionally, it receives an observation $o_t\in O$, containing the state of (partially) visible traffic participants as perceived by its sensors. 
Based on $s_t$ and $o_t$, the ego vehicle performs an action $a_t\in A$, corresponding to 
driving inputs such as velocity and steering regulation.
We assume the vehicle operates with the \textit{intent} of achieving a set of hypothesised desires $D$; we focus on short-term desires (\textit{e.g.} allowing a pedestrian to cross), but the methodology can be extended to accommodate longer-term goals (\textit{e.g.} reaching a destination). 
The hypothesised desires are defined formally at a later stage of the workflow. 
Furthermore, we assume that detailed road maps of the driving environment in which the vehicle is acting are readily available.

\subsection{State-Action Discretisation}
To generate intelligible explanations, sequences of consecutive states and actions of the ego vehicle are discretised into abstract state-action trajectories following the principles outlined by Hayes \textit{et al.}~\cite{hayes}. 
Actions are mapped from numeric driving inputs to labelled manoeuvres (\textit{e.g.} turning left, going straight). 
For states, we provide a semantically rich and interpretable representation by extracting eleven features expressed as natural language predicates (Table~\ref{tab:predicates-table}); we experimented with several state discretisers, this one was optimised for the metrics later presented, while keeping a manageable state space to allow the mapping of new observations to previously encountered states. 

These features are derived from the ego vehicle's state, observations of other traffic participants and environmental map data. They can assume one of several possible interpretable values determined and implemented in relation to the available information. 
It is paramount that these features are defined to align with the hypothesised desires motivating the ego vehicle's actions. 
\begin{table}
\caption{High-level description of predicates used to discretise the ego vehicle's state}
\label{tab:predicates-table}
\renewcommand{\arraystretch}{1.3}
\begin{tabular}[t]{ll}
\textbf{Predicate} & \textbf{Description}  \\  \hline

Velocity & Vehicle's velocity  \\  

Steering & Vehicle's steering angle \\

LanePosition &  \makecell[l]{Whether the vehicle's travel direction is aligned with its lane's\\direction, opposite to it, or if the vehicle is positioned on a divider}\\ 

BlockProgress & \makecell[l]{Vehicle's progression within its lane (\textit{e.g.} at the start, middle or \\end of the lane)}\\

NextIntersection & Vehicle's action at the upcoming intersection \\

StopAreaNearby & \makecell[l]{Presence of area where the vehicle must or may be required to \\stop}\\

CrosswalkNearby & Presence of crosswalks in the vehicle's front area \\

TrafficLightNearby& Presence of relevant traffic lights in the vehicle's front area\\

PedestrianNearby& Presence of pedestrians in the vehicle's front area\\

TwoWheelNearby& \makecell[l]{Presence of two-wheeled road users (\textit{e.g.} cyclists) in the vehicle's \\front area}\\

ObjectsNearby & Presence of relevant objects in the vehicle's front area\\ \hline
\end{tabular}
\end{table}

The discretised state-action trajectories are used to construct a policy graph as a \textit{simulacrum} of the ego vehicle's behaviour, where nodes represent states and edges correspond to actions annotated with transition probabilities between states. 
To simplify the notation, we will denote the sets of discrete states and actions as $S$ and $A$, respectively.

\subsection{Desires and Intentions}
Hypothesised desires involve executing specific actions
under certain circumstances. They are formalised and registered into the \gls{pg}. 
The definition of a desire $d \in D$ consists of two parts: the discrete state region $S_d = \{s \in S| \phi(s) \vdash d\}$ containing states that satisfy the desire's conditions, with $\phi(s)$ denoting conditions that predicates of state $s$ must meet; and $A_d \subseteq A$, the set of possible discrete actions that the vehicle should perform within $S_d$ to fulfil the desire. 
For example, let $d$ be the desire to allow a pedestrian to cross at a crosswalk; $S_d$ includes all states where a crosswalk and a pedestrian are present in the ego vehicle's front area, and $A_d$ includes actions such as braking or stopping. 
In this work, we consider several classes of desires, though other categories might be defined. These include desires associated with
high-level driving manoeuvres like lane keeping and turning
at intersections, desirable behaviours near traffic lights and
stop areas (\textit{e.g.} at stop signs), and desirable behaviours in the
proximity of vulnerable road users. A final category refers to undesirable behaviours on the road that should be avoided to ensure safety and compliance with traffic regulations.
    
    
    
    
    

The hand-crafted desires (described in Section~\ref{sec:desire_intention}) are used to compute intentions according to the algorithm formalised by Gimenez-Abalos \textit{et al.}~\cite{IPGs}. The intention value $I_d(s)$ quantifies the vehicle's intention to achieve a desire $d$ starting from state $s$, and is computed as the sum of probabilities of all possible paths originating from $s$ and reaching any state in $S_d$ where the desire is achievable and fulfilled. 
This computation is done for all defined desires in each graph state, resulting in an intention-aware \gls{pg}.

\section{Experiments} 

We evaluated the proposed methodology to generate teleological explanations of an AV's behaviour using nuScenes~\cite{nuscenes}. 

\subsection{NuScenes Dataset}
The nuScenes 
 dataset~\cite{nuscenes} consists of a collection of driving scenes captured from an AV navigating urban environments, each lasting approximately twenty seconds and consisting of forty frames sampled at 2Hz. 
 A frame includes detailed information about the state of the ego vehicle, along with annotations and states of surrounding road participants from 23 categories, including pedestrians, vehicles and generic road objects (\textit{e.g.} traffic cones). 
 The dataset also provides static map information of the driving locations, including the topology and geometry of road features such as parking areas, sidewalks, lanes, pedestrian crossings, road signs, traffic lights, lane dividers, and intersections.
We consider only scenes with available annotations (nuScenes training set)
and exclude those with incorrect pose measurements, resulting in a collection of 830 driving trajectories.

\subsection{From Driving Scenes to \gls{pg}}
The process of deriving explanations begins by representing each scene as a sequence of successive raw states and actions. 
A state contains the ego vehicle's position, heading, velocity, acceleration, steering angle, and information about nearby annotated traffic participants and road elements detected by the front camera. This includes their position, visibility (\textit{i.e.} how much the annotation is visible given the ego vehicle's 360$^o$-view) and activity status when available (\textit{e.g.} parked, moving). 
The vehicle's velocity, acceleration, and steering angle dictate actions between states.

These observations are mapped into discretised state-action trajectories and used to build the \gls{pg}. 
NuScenes lacks action labels for the ego vehicle. Therefore, we design a threshold-based heuristic to annotate actions from numeric low-level controls, which are velocity, acceleration, and steering angle. The action labels considered are: Idle, GoStraight, Gas, Brake, TurnRight, TurnLeft, and some combinations (\textit{e.g.} GasTurnRight). 
To discretise the state, we implement the predicates listed in Table~\ref{tab:predicates-table}, making extensive use of the nuScenes map:

\textbf{Velocity} can be derived from the ego vehicle's raw velocity using a threshold-based discretisation logic. Possible values are \textit{Stopped}, \textit{Slow}, \textit{Medium} and \textit{High}.

\textbf{Steering} can be derived from the ego vehicle's steering angle using a threshold-based discretisation logic. 
Possible values are \textit{Forward}, \textit{Right} and \textit{Left}.

\textbf{LanePosition} has four possible values: \textit{Aligned}, if the vehicle is travelling according to its lane's direction; \textit{Opposite}, if it is travelling against it; \textit{Centre}, if it is positioned on a lane line or road divider; and \textit{None}, if the vehicle's orientation relative to the road cannot be determined, or if it is outside the drivable area (\textit{e.g.} on a sidewalk).

\textbf{BlockProgress} 
 is quantified by dividing the lane into three equal segments, each representing a distinct traversal phase, and assigning a value accordingly: \textit{Start}, \textit{Middle} and \textit{End}. 
 If the agent is within an intersection, it is assigned \textit{Intersection}; if it is outside the drivable area, it is assigned \textit{None}.

\textbf{NextIntersection} represents the vehicle's intended course of action at the next intersection. 
Possible values are \textit{Left} (turning left), \textit{Right} (turning right), \textit{Straight} (going straight), or \textit{None} (no imminent intersection).

\textbf{StopAreaNearby} has four possible values based on the vehicle's front area: \textit{Stop}, if a stop sign is present; \textit{Yield}, if a yield sign is present; \textit{TurnStop}, for areas where the vehicle must yield to oncoming traffic when intending to make a turn; and \textit{None} otherwise. 

\textbf{CrosswalkNearby} denotes the presence of a crosswalk in
the front area of the ego vehicle (\textit{Yes/No}).

\textbf{TrafficLightNearby} denotes the presence of a traffic light
facing the ego vehicle (\textit{Yes/No}), regardless of its colour
status, as this information is not available in nuScenes’ map. 

\textbf{PedestrianNearby} is set to \textit{Yes} if the vehicle's front camera detects any object classified as pedestrians (\textit{e.g.} adults, children, police officers) within the drivable area; otherwise, the predicate is set to \textit{No}.

\textbf{TwoWheelNearby} is set to \textit{Yes} if the ego vehicle's front camera detects any object classified as motorcycles, cycles, or personal mobility vehicles with a rider and in the drivable area; otherwise, the predicate is set to \textit{No}.

\textbf{ObjectsNearby} is set to \textit{Yes} if the front camera of the ego vehicle detects elements that could affect its behaviour, and \textit{No} otherwise. 
Only detections meeting the following conditions are considered: (1) the element must be within the drivable area or a carpark area; (2) at least 60\% of the element must be visible by the 360$^o$ view of the ego vehicle; 
(3) the element must be an active (not parked) four-wheeled vehicle or an object categorised as debris, traffic cone, or pushable/pullable object (\textit{e.g.} garbage bins).

\subsection{Desires and Intentions}\label{sec:desire_intention}
In nuScenes, the ego vehicle's goal is not purely to reach a destination but rather to cruise, as the scenes are short and sometimes truncated mid-manoeuvre (\textit{e.g.} waiting in the middle of an intersection). Therefore, we define a set of \textit{safe} and \textit{unsafe} desires that are hypothesised to drive the ego vehicle's behaviour, focusing on compliance with traffic regulations and the safety of road users. \textit{Safe} desires include:
\begin{itemize}
    \item {\small\texttt{Lane Keeping}}: $S_d$ includes all states where the vehicle is moving, oriented forward, and aligned with the direction of its current lane, with no intention to turn at the next intersection. 
    $A_d$ includes going straight, accelerating or braking.
    
    \item {\small\texttt{Turn Left}}: $S_d$ includes all states where the vehicle moves at the end of a road, steers left, and plans to turn left at the next intersection. $A_d$ includes turning left, possibly combined with acceleration or deceleration. The same applies to {\small\texttt{Turn Right}}. 
    
    \item {\small\texttt{Lane Change (to lf)}}: $S_d$ includes all states where the vehicle is moving oriented toward the left and positioned
on the lane or road divider. $A_d$ includes turning left, possibly combined with acceleration, deceleration, or going straight (for longer lane changes). The same applies to {\small\texttt{Lane Change (to rt)}}.

    \item {\small\texttt{Approach Traffic Light}}: $S_d$ includes all states where the vehicle is in motion near a relevant traffic light, and $A_d$ includes stopping or braking actions. This desire should ideally consider red traffic lights only; since this information is not available in nuScenes, we assume the colour is red. The effects of this assumption are analysed in Section~\ref{intention-metrics}.
    
    \item {\small\texttt{Approach Stop Sign}}: $S_d$ includes the states where the vehicle is in motion near a stop sign without any relevant traffic lights influencing its behaviour. $A_d$ includes stopping or braking actions.
    
    \item {\small\texttt{Peds at Crosswalk}}: $S_d$ includes all states where the vehicle is travelling near a crosswalk and detects one or more pedestrians in the front. $A_d$ includes stopping or braking actions. 

    \item {\small\texttt{Non-Crosswalk Peds}}: Similar to {\small\texttt{Peds at Crosswalk}}, but the vehicle is not near a crosswalk. 
   
\end{itemize}

In the definition of \textit{unsafe} desires, $S_d$ represents a set of critical states and $A_d$ includes actions deemed unsafe or not compliant with traffic laws. These include:
\begin{itemize}
    \item {\small\texttt{Ignore Two-Wheel Vehicle}}: $S_d$ includes all states where the vehicle is travelling straight at high speed with two-wheeled vehicles ahead, and $A_d$ involves accelerating. 
    
    \item {\small\texttt{Ignore Peds (high)}}: $S_d$ includes all states where the vehicle is driving at medium or high speed and detects pedestrians in the front. $A_d$ includes all actions that do not involve braking.
    
    \item {\small\texttt{Ignore Peds (low)}}: $S_d$ includes all states where the vehicle is driving at low speed and detects pedestrians in the front. $A_d$ includes all actions involving acceleration. Differently from {\small\texttt{Ignore Peds (high)}}, actions GoStraight, TurnLeft, TurnRight are excluded from $A_d$ since, at low speeds, they may be appropriate for avoiding pedestrians.

    \item {\small\texttt{Ignore Stop Sign}}: $S_d$ includes the states where the vehicle is driving near a stop sign without any relevant traffic lights influencing its behaviour. $A_d$ includes all actions that exclude braking or stopping.
    
\end{itemize}

These desires provide a satisfying level of interpretability, as demonstrated in Section~\ref{intention-metrics}, though additional desires can be defined. Once the desires are registered in the \gls{pg}, each state's corresponding intention value is computed.

\section{Explainability Results}  

\begin{figure*}
\centering
    \includegraphics[width=\textwidth]{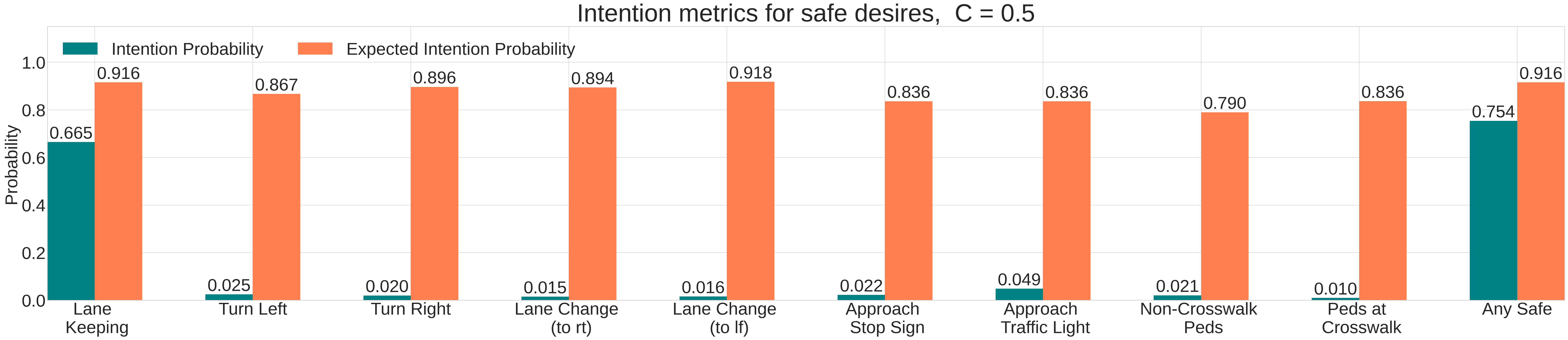}
\caption{Intention metrics for \textit{safe} desires with $C = 0.5$.
Globally, we can attribute an intention to fulfil \textit{any safe} desire to the vehicle’s behaviour $75.4\%$ of the time, with $91.6\%$ certainty that the associated desires will be fulfilled (first metrics from the right).
} 
\label{fig:good_desires_metrics}
\end{figure*}

\begin{figure}
\centering
    \includegraphics[width=0.5\textwidth]{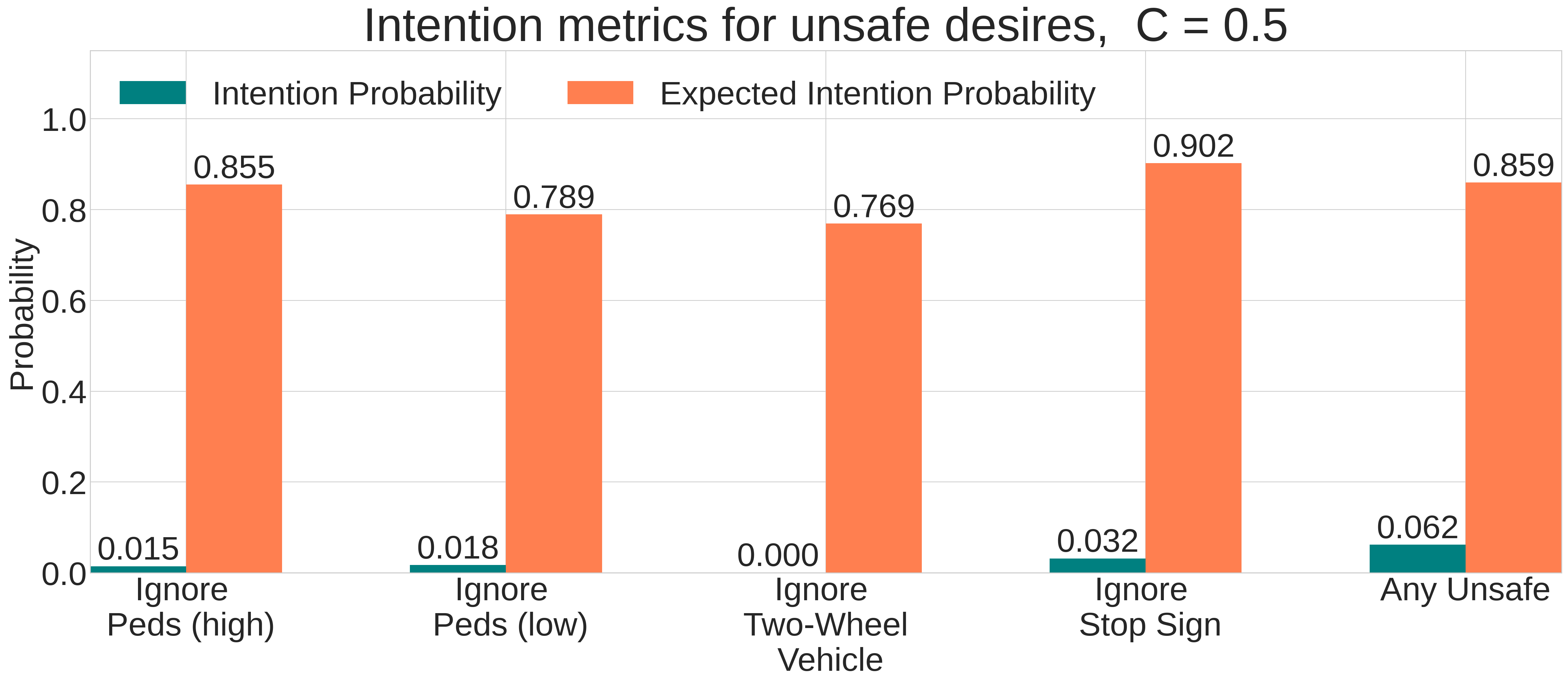}
\caption{Intention metrics for \textit{unsafe} desires with $C = 0.5$. Globally, we can attribute an intention to fulfil \textit{any unsafe} desire to the vehicle’s behaviour $6.2\%$ of the time, with $85.9\%$ certainty that associated desires will be fulfilled (first metrics from the right).
}

\label{fig:bad_desires_metrics}
\end{figure}

\subsection{Intention Metrics}\label{intention-metrics}
We assess the interpretability and reliability of the vehicle's behaviour using two metrics over a desire $d$~\cite{IPGs}: the \textit{attributed intention probability}, the probability of being in a state where the intention to fulfil $d$ is greater than a threshold $C\in(0,1]$ (interpretability), and the \textit{expected intention probability}, the probability that once attributed, an intention of $d$ is going to be fulfilled (reliability).
Higher values of $C$ increase the reliability of the explanation, as only states where the vehicle has a high probability of fulfilling the desire will be considered as states where the intention is active. Lower values of $C$ improve interpretability by attributing intentions to a broader range of states, making more of the vehicle’s behaviour interpretable. We set a commitment threshold of $C = 0.5$ to achieve a reasonable trade-off between reliability and interpretability.

Figures~\ref{fig:good_desires_metrics} and~\ref{fig:bad_desires_metrics} show intention metrics for \textit{safe} and \textit{unsafe} desires across all the ego vehicle states, providing a global overview of the agent's behaviour. 
These metrics reflect the nature of nuScenes, where the vast majority of scenes involve straightforward driving, with few instances of illegal or hazardous manoeuvres. 
We focus the analysis to the most salient desires. Except for {\small \texttt{Lane Keeping}}, the probability of manifesting intentions for other desires is generally low, reflecting their infrequent occurrence in the dataset. 
However, once attributed, these intentions are likely fulfilled, as evidenced by the consistently high expected intention probability, indicating strong reliability in the generated explanations. 

For example, the vehicle manifests the intention to allow a pedestrian to cross ({\small \texttt{Peds at Crosswalk}}) only $1.0\%$ of the time, yet when attributed, the desire is fulfilled in most cases ($83.6\%$). This approach also draws attention to the remaining $16.4\%$ of cases where the intention is not fulfilled, \textit{i.e.} the driver does not slow down or stop. 
Possible explanations include the vehicle slaloming around pedestrians, disregarding those not on its current lane, or manoeuvring around construction workers or police officers directing traffic. Introducing new values for the predicate \textit{PedestrianNearby} could help capture these distinctions. While a failure to fulfil this desire could also suggest a pedestrian being run down, this remains unlikely given the dataset. Similar results are observed when the pedestrian is not near a designated crosswalk ({\small \texttt{Non-Crosswalk Peds}}).
To further investigate potentially dangerous behaviours near pedestrians, with {\small \texttt{Ignore Peds (high)}} and  {\small \texttt{Ignore Peds (low)}}, we analyse when the vehicle intentionally ignores humans. 
These intentions are rarely attributed, 
though the high expected intention probability indicates that when attributed, they are dangerously fulfilled in most cases (85.5\% and 78.9\%, respectively). 
Intentions associated with {\small\texttt{Approach Stop Sign}} are fulfilled 83.6\% of the time; the remaining unfulfilled cases are likely due to the vehicle performing a ``rolling'' stop, where it checks for obstacles without appropriately slowing down or stopping, which could lead to collisions. 
Among \textit{unsafe} desires, {\small\texttt{Ignore Stop Sign}} is the most frequently attributed (3.2\%) and is fulfilled in 90.2\% of cases. 
For {\small\texttt{Approach Traffic Light}}, the approach allows for some global explanation to be extracted despite the lack of traffic light status. 
The desire is fulfilled $83.6\%$ of the time, while in the remaining $16.4\%$, the vehicle does not slow down or stop, likely due to a green light or, less commonly in nuScenes, a traffic violation. 

Finally, we evaluate the global interpretability and reliability of the explanations by computing the vehicle's intention to fulfil \textit{any} of the hypothesised desires~\cite{IPGs}. The metrics show that we can attribute an intention of any desire (\textit{safe} or \textit{unsafe}) to the vehicle's behaviour 75.7\% of the time, with 92.0\% certainty that the associated desires will be fulfilled. These results prove that a significant portion of the vehicle's behaviour is interpretable, as we can frequently attribute intentions behind its decisions, and that most attributed intentions lead to the fulfilment of the associated desires, reflecting the reliability of the generated explanations.

\subsection{Impact of Visibility on Driving} 
In nuScenes, scenes are collected under diverse visibility conditions, including time of the day (day or night) and weather (rainy or non-rainy). 
We analyse the impact of visibility on the vehicle's behaviour by constructing separate \gls{ipg}s on the night, daytime, rainy, and non-rainy scenes and comparing intention metrics over the hypothesised desires. 
The metrics show that stop sign desires are particularly affected by nighttime conditions (Table~\ref{tab:low-visibility-table}): at night, unlike during the day, the vehicle never manifests the intention to approach or ignore a stop sign. 
These results may uncover limitations in the vehicle's visual system, inaccuracies in human-annotated traffic sign labels within the nuScenes dataset, or biases in the collection of nighttime scenes, which may have been gathered in regions lacking stop signs. 

\begin{table}
    \centering
    \caption{Intention metrics for stop sign desires under varying visibility conditions, with $C = 0.5$. \textit{A.I.} and \textit{E.I.} represent the attributed and expected intention probabilities.}
    \label{tab:low-visibility-table}

\addtolength{\tabcolsep}{+0.4em}

    \begin{tabular}{lcccc|cccc}
        & \multicolumn{2}{c}{\textbf{Day}} & \multicolumn{2}{c|}{\textbf{Night}} & \multicolumn{2}{c}{\textbf{No Rain}} & \multicolumn{2}{c}{\textbf{Rain}} \\
        \cmidrule(lr){2-9} 
        \textbf{Desire} & \textit{A.I.} & \textit{E.I.} & \textit{A.I.} & \textit{E.I.} & \textit{A.I.} & \textit{E.I.} & \textit{A.I.} & \textit{E.I.} \\
        \midrule
        Approach Stop Sign & 0.026 & 0.836 & \textbf{0} & \textbf{0} & 0.023 & 0.843 & 0.020 & 0.843 \\
        Ignore Stop Sign & 0.036 & 0.901 & \textbf{0} & \textbf{0} & 0.032 & 0.901 & 0.033 & 0.895  \\ \hline
    \end{tabular}
\end{table}

\subsection{Scene Analysis and Question Answering} 

\begin{figure}[t]
\centering
    \includegraphics[width=0.75\textwidth]{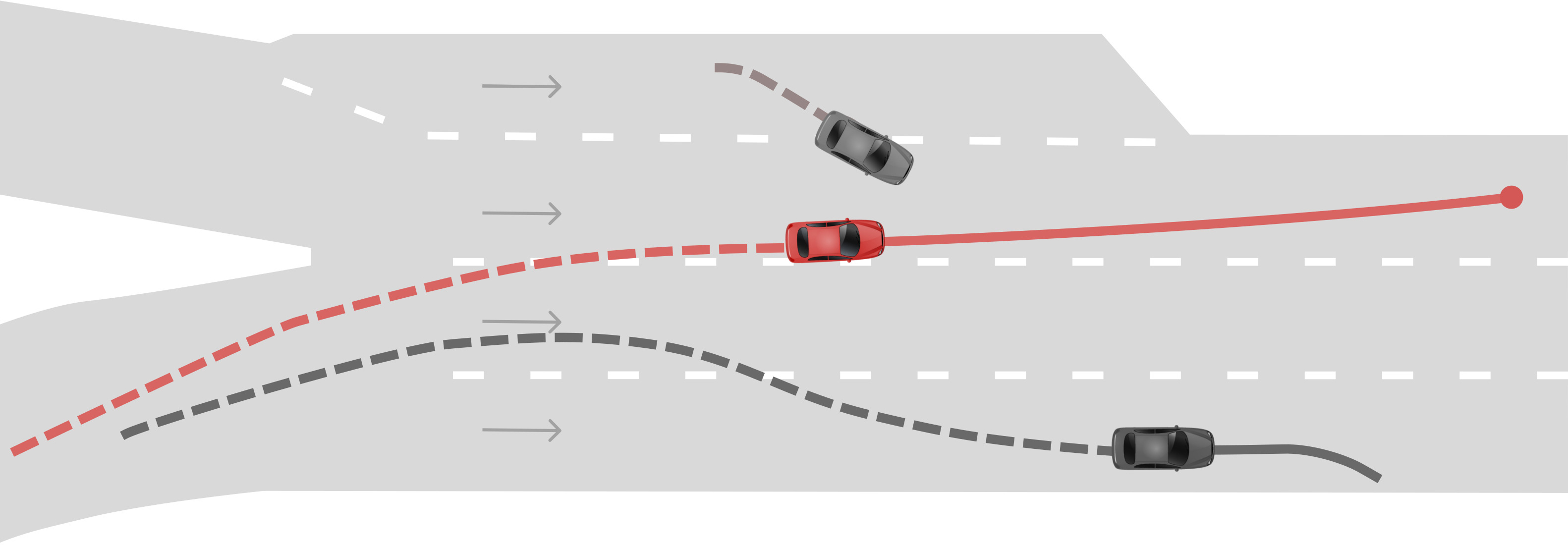}
\caption{Abstract representation of scene 1084 of nuScenes. The ego vehicle is represented in red, with its final state marked with a red dot. Non-ego vehicles are represented in grey. The scene is captured at state $s = 20$, when the ego vehicle is nearly stationary in reaction to the abrupt lane intrusion of another vehicle.} 
\label{fig:scene-example}
\end{figure}

\begin{figure}
\centering
    \includegraphics[width=1\textwidth]{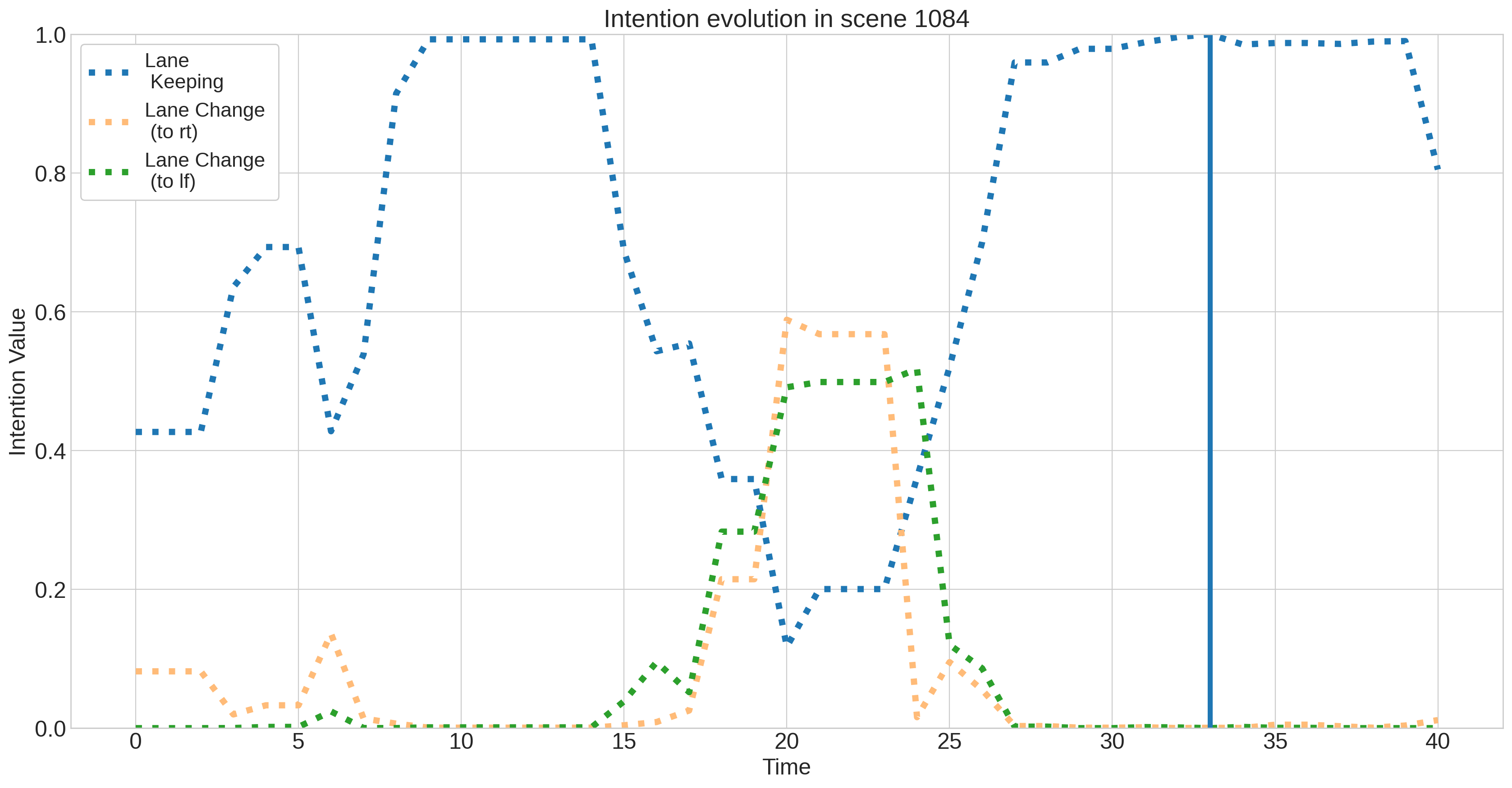}
\caption{Temporal evolution of the ego vehicle's intention at each state in scene 1084, for different desires. For interpretability purposes, only desires that attain an intention value greater than 0.2 at least once throughout the scene are included in the visualisation. Vertical lines in the plot refer to the fulfilment of desires of the corresponding colour.} 
\label{fig:int-progression}
\end{figure}

Finally, we demonstrate the potential of our approach in extracting local explanations from a specific driving scenario.
Figure~\ref{fig:scene-example} illustrates a high-level representation of the analysed scene\footnote{Scene available at \url{https://www.nuscenes.org/nuscenes?sceneId=scene-1084}.}, in which the ego vehicle merges onto a main road from a secondary lane and subsequently transitions to the middle-left lane of the road. As the lane changing manoeuvre nears completion, a non-ego vehicle merges from the far-left lane into the ego vehicle's trajectory without signalling, forcing the ego vehicle to brake. The non-ego vehicle also decelerates, and pauses its attempt to merge
. Once the situation stabilises, the ego vehicle resumes its trajectory, completing the lane change and maintaining a straight path. 

An initial explanation is derived by examining how the ego vehicle's intentions change over time. The temporal evolution of the ego vehicle's intentions throughout the scene is shown in Figure~\ref{fig:int-progression}. The agent is initially committed to lane keeping, with a notable shift in intentions between $s=14$ and $s=27$. Here, the ego vehicle first exhibits an increasing intention toward a leftward lane change, which is subsequently overtaken by a stronger intention to change lane to the right, likely as a response to the sudden incursion of the non-ego vehicle. After mitigating the risk of a potential collision, the ego vehicle reaffirms its commitment to lane keeping and stabilises its behaviour.
This analysis reveals unintentional or anomalous behaviours, exposes fulfilled and unfulfilled desires, and facilitates the formulation of hypotheses about the underlying reasoning of the vehicle. By comparing the actual scene against the temporal intention evolution, a discrepancy emerges between the observed behaviour and the computed intentions: in the raw scene, the ego vehicle starts executing a leftward lane change shortly after merging into the main road, whereas the intention evolution plot suggests that the intention of this desire is manifested later in time. Upon further investigation, we traced this anomaly to an inconsistency in the nuScenes map: the dataset map does not represent a portion of lane divider that is physically present in the real-world scene, and this omission affected the discretisation of the ego vehicle's initial states in the scene, leading to a misalignment in the computed intentions. 

To further elucidate the motivational state of the ego vehicle, the computed intentions are used to answer three explainability questions, formalised by Gimenez-Abalos et \textit{al.}~\cite{IPGs}: 

\begin{itemize}
    \item \textit{``What do you intend to do in state \textit{s}?''}, obtained by returning all desires with  $I_d(s) \geq C$. We set $C=0.5$ to have a reasonable trade-off between reliability and interpretability.
    
    \item \textit{``Why would you do action $a$ in state $s$?''}, answered by determining whether $a$ results in an expected increase in the intention with a particular desire, or if it has some probability of an expected increase in intention for that desire. If $s$ has no attributed intention or if $a$ has no acceptable metric for any desire, then the action is considered unintentional with respect to the \gls{pg} and the hypothesised set of desires.
    
    \item \textit{``How do you plan to fulfil intention \textit{$I_d$} in state \textit{s}?''}, obtained by retrieving the optimal path, based on intention values, from $s$ to a state in $S_d$.
\end{itemize}

Having identified a critical region through the previous visual explanation, we formulate explainability queries at two specific states: $s_1 = 14$, preceding the shift in intentions, and $s_2 = 20$, where the ego vehicle is nearly halted due to a non-ego vehicle abruptly merging into its lane. We compare the responses at these two states to assess how the ego vehicle's intentions and its plans to fulfil them vary in response to environmental changes.

At $s_1 = 14$, the ego vehicle is observed travelling at high speed with a forward orientation, with no nearby road signs or vulnerable road users, while detecting one or more objects ahead. The answer to \textit{what} the ego vehicle intents to do is strongly commit (0.993) to {\small\texttt{Lane Keeping}}. The subsequent query, about \textit{how} the agent plans to achieve this intent, outputs the set of actions in $A_{lane\_keeping}$: Brake, Gas, or GoStraight. Given that the intention value is already near its maximum, the execution of a single action by the ego vehicle is sufficient to approach desire fulfilment, requiring no further adjustments to its trajectory. On the other hand, at state $s_2 = 20$ (illustrated in Figure~\ref{fig:scene-example}) the scenario evolves: the ego vehicle is positioned on a lane divider and travelling at low speed, still oriented forward and without nearby traffic signs or vulnerable users, while detecting one or more objects ahead. The response to \textit{what} the ego vehicle intends to do in $s_2$ involves the desire to {\small\texttt{Lane Change (to rt)}} (0.589). 
The intention of {\small\texttt{Lane Keeping}} falls below the commitment threshold (as shown in Figure~\ref{fig:int-progression}), and querying \textit{how} this intention might be fulfilled reasonably yields a more complex plan than the one computed at $s_1$, as detailed in Table~\ref{tab:QA-lanekeeping-table}.

\begin{table*}[t]
\centering
\caption{Response to ``\textit{How do you plan to fulfil lane keeping from $s_2$}?'' as a sequence of steps, each including the action the vehicle would execute in the current state and how it believes the state could change after performing the action. The sequence starts at $s_2$ and progresses towards a state in $S_{lane\_keeping}$, where the desire is fulfilled through an action selected from $A_{lane\_keeping}$.
}
\label{tab:QA-lanekeeping-table}
\begin{tabular}{lll}
\textbf{Action} & \textbf{State Transition} & \textbf{$I_d(s)$}   \\  \hline

GasTurnRight & \makecell[l]{Remove: Steering(\textit{Forward}) \\ Add: Steering(\textit{Right})} &  0.209 \\  \hline

GasTurnRight & \makecell[l]{Remove: BlockProgress(\textit{Middle}) \\ Add: BlockProgress(\textit{Start})} &  0.486\\ \hline

Gas & \makecell[l]{Remove:
Steering(\textit{Right}) \\ Add: Steering(\textit{Forward})} &  0.486 \\  \hline

GasTurnRight & \makecell[l]{Remove: FrontObject(\textit{Yes}), Velocity(\textit{Low}), Steering(\textit{Forward}) \\ Add: FrontObject(\textit{No}), Velocity(\textit{Medium}), Steering(\textit{Right})  } &  0.551\\ \hline

TurnRight & \makecell[l]{Remove: StopAreaNearby(\textit{No}) \\ Add: StopAreaNearby(\textit{TurnStop})} &  1.0\\ \hline

Brake & \makecell[l]{Remove: Steering(\textit{Right}), LanePosition(\textit{Centre})\\ Add: Steering(\textit{Forward}), LanePosition(\textit{Aligned}) } &  1.0\\ \hline

\end{tabular}
\end{table*}

\begin{table*}[t]
\centering
\caption{Response to ``\textit{How do you plan to fulfil lane change (to rt) from $s_2$}?''. The short plan suggests that the desire is readily attainable from $s_2$.} 
\label{tab:QA-lane-change-table}
\begin{tabular}{lll}
\textbf{Action} & \textbf{State Transition} & \textbf{$I_d(s)$}   \\  \hline
GasTurnRight & \makecell[l]{Remove: Steering(\textit{Forward}) \\ Add: Steering(\textit{Right})} &  1.0 \\  \hline \\
\end{tabular}
\end{table*}



We further query \textit{how} the ego vehicle plans to fulfil its lane-changing intention from $s_2$. The response provides an optimal plan 
(Table~\ref{tab:QA-lane-change-table}); however, in the actual scene, the agent does not execute the action outlined in this plan to advance its intention; instead, it opts to break. To understand the motivations behind this behaviour, we ask \textit{why} the ego vehicle chooses to Brake at $s_2$, as well as \textit{why} it would have performed GasTurnRight, the computed optimal action for achieving {\small\texttt{Lane Change (to rt)}}. 
As outlined in Table~\ref{tab:QA-actions-table}, braking is associated with a negligible expected intention increase towards {\small\texttt{Lane Change (to lf)}}, which is weaker than the increase towards {\small\texttt{Lane Change (to rt)}} expected from executing GasTurnRight. Although decelerating may appear sub-optimal, 
it is likely intended to favour other desires (\textit{e.g.} obstacle avoidance, emergency braking) not currently represented within the hypothesised desire set $D$. Expanding $D$ to incorporate these additional desires further improves the interpretability of the vehicle's behaviour and will be explored in future work.

\begin{table*}[t]
\centering
\caption{Answers to ``\textit{Why would you take action $a$ in $s_2$}?'' for a set of actions
}
\label{tab:QA-actions-table}
\begin{tabular}{ll}
\textbf{Action} & \textbf{Answer} \\  \hline

Brake & \makecell[l]{
For the purpose of furthering Lane Change (to lf), as it has a 0.200\\probability of an expected intention increase of 0.008}\\  \hline

GasTurnRight & \makecell[l]{
For the purpose of furthering Lane Change (to rt) as it has a 0.200\\probability of an expected intention increase of 0.411}\\ \hline


\end{tabular}
\end{table*}

\section{Discussion and Future Work}
This paper introduces a \textit{post-hoc}, model-agnostic explainability method designed to provide interpretable and reliable teleological explanations for autonomous vehicle behaviour in urban environments. Our approach, rooted in Intention-aware Policy Graphs, offers intelligible global and local explanations of the vehicle's goals and the corresponding plans to achieve them. It enables the distinction between intentional and unintentional behaviour, as well as fulfilled and unfulfilled desires, while also identifying unsafe driving practices, thereby enhancing transparency and accountability in AVs. 
The findings for stop sign desires under \textit{low} visibility conditions and the emerged inconsistencies in the nuScenes map demonstrate the method's effectiveness in revealing possible vulnerabilities in autonomous driving datasets or models. 
The limitations of this work include the dataset used, given its limited size and focus on basic driving scenarios, in addition to extensive domain knowledge when defining state discretisation and vehicle desires.
Future research will include a user study to assess the impact of the generated explanations on trust among non-expert users. Additionally, we plan to expand the framework to account for multi-agent interactions, incorporating desires and intentions of other traffic participants.

\begin{credits} 
\subsubsection{\ackname} This work is partially funded by the European Commission through the AI4CCAM project (Trustworthy AI for Connected, Cooperative Automated Mobility) under grant agreement No 101076911. Views and opinions expressed are, however, those of authors only and do not necessarily reflect those of the European Union or CINEA. Neither the European Union nor the granting authority can be held responsible for them. Furthermore, Sara Montese is supported by the fellowship within the ``Generación D'' initiative, \href{https://www.red.es/es}{Red.es}, Ministerio para la Transformación Digital y de la Función Pública, for talent attraction (C005/24-ED CV1). Funded by the European Union NextGenerationEU funds, through PRTR.

\subsubsection{\discintname} The authors have no competing interests to declare that are relevant to the content of this article.
\end{credits}

%
%
%
\printbibliography

\end{document}